\PassOptionsToPackage{table}{xcolor}
\documentclass[pdflatex,sn-mathphys-num]{sn-jnl}

\usepackage{graphicx}%
\usepackage{multirow}%
\usepackage{amsmath,amssymb,amsfonts}%
\usepackage{amsthm}%
\usepackage{mathrsfs}%
\usepackage[title]{appendix}%
\usepackage{xcolor}%
\usepackage{textcomp}%
\usepackage{manyfoot}%
\usepackage{booktabs}%
\usepackage{algorithm}%
\usepackage{algorithmicx}%
\usepackage{algpseudocode}%
\usepackage{listings}%
\usepackage{subcaption}

\theoremstyle{thmstyleone}%
\theoremstyle{thmstyletwo}%

\theoremstyle{thmstylethree}%
\definecolor{mygray}{gray}{0.7}
\begin{document}

\title[Article Title]{Cooperative Hybrid Multi-Agent Pathfinding Based on Shared
Exploration Maps}

\author*[1]{\fnm{Ning} \sur{Liu}}\email{ning.liu@research.uwa.edu.au}

\author[2]{\fnm{Sen} \sur{Shen}}\email{senshen@cuhk.edu.hk}

\author[1]{\fnm{Xiangrui} \sur{Kong}}\email{xiangrui.kong@research.uwa.edu.au}

\author[1]{\fnm{Hongtao} \sur{Zhang}}\email{hongtao.zhang@research.uwa.edu.au}

\author[1]{\fnm{Thomas} \sur{Bräunl}}\email{thomas.braunl@uwa.edu.au}

\affil*[1]{\orgdiv{Department of Electrical, Electronic and Computer Engineering}, 
\orgname{The University of Western Australia}, 
\orgaddress{\city{Perth}, \postcode{6009}, \state{WA}, \country{Australia}}}

\affil[2]{\orgdiv{Department of Mechanical and Automation Engineering}, 
\orgname{The Chinese University of Hong Kong}, 
\orgaddress{\city{Shatin}, \postcode{999077}, \country{Hong Kong SAR}}}


\abstract{Multi-Agent Pathfinding is used in areas including multi-robot formations, warehouse logistics, and intelligent vehicles. However, many environments are incomplete or frequently change, making it difficult for standard centralized planning or pure reinforcement learning to maintain both global solution quality and local flexibility. This paper introduces a hybrid framework that integrates D* Lite global search with multi-agent reinforcement learning, using a switching mechanism and a freeze-prevention strategy to handle dynamic conditions and crowded settings. We evaluate the framework in the discrete POGEMA environment and compare it with baseline methods. Experimental outcomes indicate that the proposed framework substantially improves success rate, collision rate, and path efficiency. The model is further tested on the EyeSim platform, where it maintains feasible Pathfinding under frequent changes and large-scale robot deployments.}


\maketitle

\section{Introduction}
Multi-Agent Pathfinding (MAPF) is an important topic in robotics and artificial intelligence. It aims to enable multiple agents to perform movement tasks while minimizing conflicts \cite{sharon2015conflict,chen2023transformer,chandarana2021planning}. With the increasing demand for unmanned aerial vehicle formations \cite{sinnemann2022systematic}, warehouse logistics robots \cite{thummalapeta2023survey}, and autonomous vehicles \cite{xiao2024artificial, han2024deep}, multi-agent systems have grown in scale, number of agents, and interaction complexity \cite{sun2025multi}. These scenarios often include unknown or dynamic obstacles \cite{baglioni2024novel}, and traditional centralized planning methods that depend on complete prior knowledge do not always provide real-time and adaptive performance \cite{mahmoudzadeh2024holistic}.

In recent years, various methods have been introduced to address MAPF. Early work primarily employed centralized planning approaches, such as A* \cite{hart1968formal} and its extensions, which can provide high-quality paths in small or static environments. Moreover, deep reinforcement learning (DRL) and multi-agent reinforcement learning (MARL) \cite{arulkumaran2017deep,ma2024efficient} have demonstrated the ability to handle more complex situations \cite{antonyshyn2024deep,xu2022autonomous}. For instance, PRIMAL \cite{sartoretti2019primal} combines reinforcement learning and imitation learning to train fully decentralized policies, allowing agents to make autonomous decisions with implicit coordination in partially observable environments. However, dedicated research focusing on partially observable scenarios remains limited. Many learning-based approaches still rely on global information for training and decision-making, and when agents must travel long distances over an extended period, reinforcement learning can encounter adaptability problems, especially if the environment changes frequently.

Despite these advancements, several challenges remain. Centralized techniques such as A* are susceptible to computational bottlenecks when many agents or rapidly changing conditions are involved \cite{li2019improved,grenouilleau2019multi}. On the other hand, distributed algorithms may struggle to resolve conflicts under strict local constraints or strong agent coupling \cite{rizk2018decision}. DRL and MARL can learn policies in partially observable settings, but they frequently require large training datasets and extended training time \cite{orr2023multi}, and frequent changes in the environment can impact both convergence and policy stability \cite{9340876}. Hence, balancing adaptability and computational efficiency remains an important concern in the MAPF field \cite{zhu2024survey}.

\begin{figure*}[t]
    \centering
    \includegraphics[height=2.5in]{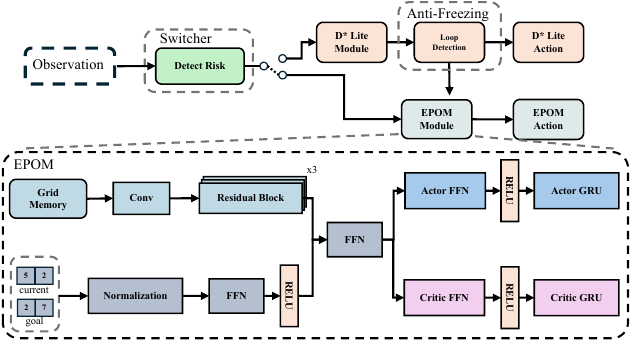}
    \caption{
Framework of CHS. The system detects loops to trigger reinforcement learning actions, or uses a switching mechanism to select between D* Lite and EPOM based on agent density. The EPOM module processes observations through a ResNet encoder and GRU networks. This architecture enables dynamic switching between global and local planning while maintaining path consistency.
    }
    \label{fig:framework}
\end{figure*}

To address these issues, this work proposes a hybrid approach Cooperative Hybrid Multi-Agent Pathfinding Based on Shared Exploration Map (CHS) as shown in Fig.~\ref{fig:framework} that integrates global search with MARL for large-scale and dynamic environments. Our method combines D* Lite's incremental planning capabilities with MARL's adaptive local decision-making to maintain both global path optimality and local flexibility. The system begins by detecting potential loops and congestion in the environment. When such conditions are detected, it activates the appropriate decision-making module—either global D* Lite for efficient pathfinding or local MARL for collision avoidance in crowded areas. Meanwhile, agents continuously update and share only incremental environmental changes, reducing redundant planning while maintaining comprehensive spatial awareness. We validate our approach in both the POGEMA discrete environment \cite{skrynnik2024pogema} and the EyeSim physical platform\cite{Braeunl2023}, demonstrating its effectiveness in various scenarios.

Our contributions can be summarised as:
\begin{enumerate}
    \item We propose a novel hybrid framework that integrates D* Lite global search with MARL, leveraging the incremental update capabilities of D* Lite to eliminate complete re-planning when environmental conditions change.
    \item We develop a shared exploration map mechanism through which agents exchange only incremental environmental modifications, substantially reducing communication overhead while preserving comprehensive spatial awareness for coordinated decision-making.
    \item We empirically evaluate our approach through extensive experiments in both POGEMA simulation environment and EyeSim physical platform, demonstrating significant improvements in success rate, collision avoidance, and path optimization, particularly in scenarios characterized by large-scale agent deployments and frequent environmental perturbations.
\end{enumerate}

\section{Related Work}

\subsection{Traditional Search Methods}
Early approaches to multi-agent path planning relied on centralized methods such as Conflict-Based Search and Improved Conflict-Based Search \cite{boyarski2015icbs}. These methods separate conflict detection from conflict resolution, which can produce high-quality paths in small-scale or static environments. However, when the number of agents or moving obstacles grows, the computational cost increases significantly \cite{li2019multi}. Priority-based strategies reduce collisions through a hierarchy of agent priorities but can lead to deadlocks when passages are narrow or communication is restricted \cite{ferner2013odrm}. To handle larger or partially unknown layouts, incremental methods like D* Lite \cite{al2011d} update existing paths instead of re-planning from scratch, and Distributed RRT \cite{desaraju2011decentralized} constructs sampling trees among agents to reduce centralized demands. Although these methods distribute planning across agents, spatial-temporal coupling remains challenging in complex maps, and agents may become stuck if their local routes interfere \cite{bagad2024optimizing}.

\subsection{Learning-based Methods}
MARL methods address uncertainties through adaptive strategies that emerge from local observations. For instance, PRIMAL and its follow-up PRIMAL2 \cite{damani2021primal} use DRL in grid-based environments, but collisions may still occur under heavy congestion. PICO \cite{li2022multi} applies graph neural networks for local communication, achieving higher success rates on large-scale maps, although global coordination is not guaranteed. Some hybrid methods combine classic planning with local DRL. FOLLOWER \cite{skrynnik2024learn} first assigns global paths to each agent and then uses learning-based adjustments for local decisions, which reduces collisions and enhances path quality. Another hybrid system \cite{ye2023toward} uses D* Lite for global pathfinding and a multi-agent deep deterministic policy gradient module for local control, improving energy efficiency and obstacle avoidance. Nevertheless, these methods depend on complete map information and large training sets, making them less robust when the environment or agent count changes, which restricts their scalability and generalization in real-world settings.

\subsection{Communication Mechanisms}
In partially observable or highly dynamic environments, communication serves as a critical mechanism for reducing uncertainty. While some approaches employ broadcast strategies \cite{shaw2022formic}, such methods introduce significant bandwidth consumption and latency issues as network scale increases \cite{hong2019investigating}. Various solutions have emerged to address these limitations, including transmission range constraints and adaptive update frequencies based on spatial-temporal factors \cite{ma2021learning,ma2021distributed,li2021message}. Alternative approaches leverage graph-based communication protocols that selectively filter and transmit only essential information \cite{tang2024ensembling}. Despite these advancements, the optimizing observation sharing remains a fundamental challenge in multi-agent systems. Suboptimal communication strategies can significantly degrade the performance of otherwise well-designed planning frameworks, resulting in network congestion or insufficient distribution of critical environmental data.

\section{HYBRID POLICY}

\subsection{Problem Formulation}

We define the partially observable MAPF problem on an undirected graph \(G=(V,E)\) with a set of \(n\) agents \(A = \{1,2,\ldots,n\}\). Each agent \(i \in A\) is assigned a start vertex \(s_i \in V\) and a goal vertex \(g_i \in V\). An agent may either wait in its current vertex or move to an adjacent vertex at each discrete time step. A path for agent \(i\) can be represented by a function \(\pi_i : \{0,1,\ldots,T_i\}\to V, \pi_i(0) = s_i, \pi_i(T_i) = g_i\), where \(T_i\) is the time at which agent \(i\) reaches its goal.

Collision-free movement requires \(\pi_i(t)\neq \pi_j(t)\) for all \(i\neq j, t\) to avoid occupying the same vertex at the same time, and \((\pi_i(t),\pi_i(t+1))\neq (\pi_j(t+1),\pi_j(t))\) for all \(i\neq j, t\) to prevent two agents from traversing the same edge in opposite directions at the same time.

In the PO-MAPF setting, each agent only has access to local observations rather than full information about the environment. Specifically, an agent at vertex \(v\) at time \(t\) receives local information \(\mathcal{O}(v,t)\) via \(\mathcal{O}: V \times \mathbb{N} \to \mathcal{X}\), and lacks details about other agents' goals \(\{g_j\}_{j \neq i}\) or their planned paths \(\{\pi_j\}_{j \neq i}\). The objective is to plan paths for all agents to reach their goals while respecting collision constraints, despite only having local observations.

\subsection{Multi-agent APPO with Grid Memory}

Proximal Policy Optimization \cite{schulman2017proximal} is known for its stable performance in single-agent reinforcement learning. However, when multiple agents must gather trajectories and update policies in parallel, communication overhead becomes a challenge. Asynchronous Proximal Policy Optimization (APPO) \cite{yu2022surprising} addresses this by allowing each agent to collect trajectories and update its policy parameters in parallel, reducing overhead and speeding up training.

In our framework, each agent's action at time $t$ depends on its local observation $o^i_t$ and hidden state $h^i_{t-1}$. Once enough trajectories are gathered, each agent updates its policy asynchronously. The actor picks actions using $\pi(a \mid s;\theta)$, while the critic provides value estimates $V(s;\phi)$ or $\hat{V}(o^i_t,h^i_{t-1};\phi^i)$, offering feedback for policy improvement. By tuning hyperparameters such as the discount factor $\gamma$, learning rate, and entropy coefficient $\beta$, the method achieves both stability and efficiency.

In partially observable MAPF, let $N$ agents each observe only local obstacles and agent positions, as well as its own and goal coordinates. A recurrent neural network (RNN) maintains a hidden state $h^i_{t}$:
\begin{equation}
    h^i_t = f\bigl(h^i_{t-1}, o^i_t\bigr),
\end{equation}
aggregating previous observations. During sampling, each agent collects trajectories $\tau^i = \bigl\{(o^i_t, a^i_t, r^i_t)\bigr\}_{t=0}^T$. Let $\theta^i_{\text{old}}$ be the previous parameters; then the probability ratio is
\begin{equation}
    r^i_t(\theta^i) 
    = 
    \frac{\pi^i\bigl(a^i_t \mid o^i_t, h^i_{t-1}; \theta^i\bigr)}
    {\pi^i\bigl(a^i_t \mid o^i_t, h^i_{t-1}; \theta^i_{\text{old}}\bigr)},
\end{equation}
and a clipping mechanism limits large policy shifts. Generalized advantage estimation (GAE) can further stabilize learning with
\begin{equation}
    L^i(\theta^i, \phi^i) 
    = 
    L_{\text{clip}}^i(\theta^i)
    - 
    c_1\, L_{\text{v}}^i(\phi^i)
    + 
    c_2\, L_{\text{entropy}}^i(\theta^i).
\end{equation}

\begin{table}[htbp]
\centering
\caption{Reward Function Design}
\label{tab:reward_function}
\begin{tabular}{l l p{6.5cm}}
\hline
\textbf{Condition} & \textbf{Reward} & \textbf{Explanation} \\
\hline
Every time step & -0.0001 & Penalizes excessive resource usage over time \\
Collision & -0.0002 & Discourages collisions with obstacles or agents \\
Reaching the goal & 1.0 & Encourages timely task completion \\
\hline
\end{tabular}
\end{table}
On top of APPO, EPOM~\cite{skrynnik2023switch} introduces a grid memory for managing obstacle information. Unlike standard APPO, this grid memory is updated alongside the RNN hidden state and stores incremental obstacle data from the global map $M$. It scales to different map sizes and ensures agents do not rely on outdated obstacle information. 
Once the grid memory has recorded new obstacle information, the agent can include these blocked locations in its decision-making at the next time step, thus avoiding outdated environmental data.

Building on this, and to balance forward efficiency and safety, we design a reward function (summarized in Table \ref{tab:reward_function}) that includes a time-step penalty, a collision penalty, and a positive reward for reaching the goal. The small negative reward at each time step discourages excessive resource usage, and the collision penalty helps the agent avoid colliding with obstacles or other agents. Meanwhile, the positive reward upon reaching the goal encourages timely task completion. By integrating the grid memory with the reward function, the agent remains sensitive to the latest obstacle information and can balance local decisions with global objectives.

\subsection{Switching Mechanism and Anti-Freezing Strategy}

\begin{algorithm}[ht]
\caption{Switching Mechanism and Anti-Freezing Strategy (Range-Based)}
\label{alg:switch_anti_freeze}
\begin{algorithmic}[1]

\Require Agent \(i\) with position \(\mathbf{x}^i\)
\Require Local observation \(o^i\)
\Require Position history \(\mathit{hist}_t\)
\Require Global map \(\mathit{map}_t\)

\Ensure \(\mathit{map}_{t+1}\)
\Ensure \(\mathit{hist}_{t+1}\)

\State \(n \gets \operatorname{Count}\bigl(\mathbf{x}^i,\, o^i\bigr)\)

\If{\(n > 4\)}
    \State \(\mathrm{mode}_i \gets \mathrm{LOCAL\_RL}\)
\Else
    \State \(\mathrm{mode}_i \gets \mathrm{D*Lite}\)
\EndIf

\If{\(\mathrm{mode}_i = \mathrm{LOCAL\_RL}\)}
    \State \(a^i \gets \mathrm{RL\_Module}\bigl(\mathbf{x}^i,\, o^i\bigr)\)
\Else
    \State \(a^i \gets \mathrm{D*Lite}\bigl(\mathbf{x}^i,\, o^i,\, \mathit{map}_t\bigr)\)
    
\EndIf

\State \textbf{Execute} \(a^i\) and observe new position \(\mathbf{x}^i_{\mathrm{new}}\)
\State \(\mathit{map}_{t+1} \gets \operatorname{MapUpdate}\bigl(\mathit{map}_t,\, o^i\bigr)\)
\State \(\mathit{hist}_{t+1} \gets \mathit{hist}_t \cup \{\mathbf{x}^i_{\mathrm{new}}\}\)

\State \Return \(\mathit{map}_{t+1},\, \mathit{hist}_{t+1}\)

\end{algorithmic}
\end{algorithm}

To illustrate how these components are combined in our method, we provide an overview in Fig.~\ref{fig:framework}. As shown, the system checks for loops, switches between local reinforcement learning and a global D* Lite planner, and processes extended observations via EPOM. The subsequent sections elaborate on each module’s role in addressing partially observable MAPF.

A key motivation for this design is that a single global planner, such as D* Lite, can become less effective when agents are closely packed or when local obstructions frequently appear. In particular, if an agent repeatedly revisits the same region, it may not make adequate progress toward its goal \cite{mao2024multi}. To address these situations, we introduce a simple rule that switches the agent to a local RL approach when the risk of congestion is high or when loop behavior is detected in D* Lite. This hybrid design reduces the risk of repeated pathing in congested areas and ensures that each agent can still maintain a global route while responding immediately to local conditions.

Algorithm~\ref{alg:switch_anti_freeze} illustrates how each agent chooses between using the D* Lite global planner or a local RL module. If the agent finds more than four neighbors, it activates the RL module to handle tighter collision avoidance. Otherwise, it continues with the D* Lite path. This ensures that when the agent is in a crowded area, it can rely on local decisions that quickly adjust to unpredictable movements.

We also include an anti-freezing strategy in D* Lite so that agents do not get stuck by moving back and forth or waiting indefinitely \cite{wang2020walk}. If an agent's position matches its last or second-to-last location, it is considered a loop \cite{okumura2022priority}. The agent then switches to local RL to break out of the loop and keep moving toward its destination. This simple check avoids repeated back-and-forth motions in narrow passageways and helps ensure overall forward movement.

\begin{algorithm}[h]
\caption{D* Lite (Loop Detection)}
\label{alg:dslite}
\begin{algorithmic}[1]

\Require \(o^i, \;\mathit{map}_t, \;\mathbf{x}^i, \;\mathit{hist}_t\)
\Ensure \(a^i\)

\State \(\mathrm{plan} \gets \mathrm{D*Lite}\bigl(\mathit{map}_t,\, \mathbf{x}^i,\, o^i\bigr)\)

\If{\(\mathrm{plan} \neq \varnothing\)}
    \State \(a^i \gets \operatorname{GetFirstAction}\bigl(\mathrm{plan}\bigr)\)
\Else
    \State \(a^i \gets \operatorname{Replan}\bigl(\mathbf{x}^i,\, \mathit{map}_t\bigr)\)
\EndIf

\If{\(\operatorname{DetectLoop}\bigl(\mathbf{x}^i,\, \mathit{hist}_t\bigr)\) \textbf{ or } \(a^i = \mathrm{None}\)}
    \State \(a^i \gets \mathrm{RL\_Module}\bigl(\mathbf{x}^i,\, o^i\bigr)\)
\EndIf

\State \Return \(a^i\)

\end{algorithmic}
\end{algorithm}

The switching mechanism employs a simple threshold to gauge congestion. Let \(n\) 
be the set of agents within a predefined range of agent \(i\). If \(\lvert n \rvert > 4\), agent \(i\) suspends
its D* Lite plan and switches to local RL. This straightforward rule provides a clear
condition for switching without relying on complex crowding metrics.

To detect loops or freezing, each agent records its recent positions. Let $\mathbf{x}^i_t$ be the agent's location at time~$t$. If $\mathbf{x}^i_t$ is the same as $\mathbf{x}^i_{t-2}$, we classify this as loop behavior. At that moment, the agent relies on an RL-based action to prevent repeated navigation in the same spot.

Algorithm~\ref{alg:dslite} demonstrates how D* Lite can be modified to check for loops. If the agent is stuck or has no valid action, it resorts to an RL-based movement command. This loop detection step helps the agent avoid repeatedly occupying the same cells and allows it to switch to a more flexible local planner whenever necessary \cite{luna2011push}.

\subsection{Construction of the Shared Map and Grid Memory}

Maintaining a fully synchronized global map in a distributed setting can lead to high communication overhead. Let \(M^{(t)}\) be the global map at time \(t\). To keep track of each agent's local observations, we introduce a shared map module that aggregates new data from multiple agents. Suppose agent \(i\) discovers additional information at time \(t\) regarding free or blocked locations (which can be vertices or edges). We denote these local updates $\Delta m^i_t$ as
\[
\Delta m^i_t \;=\; \{\, (x,\, \mathrm{val}) \mid x \in \mathcal{X}^i_t,\;\mathrm{val} \in \{0,1\} \},
\]
where \(\mathcal{X}^i_t\) is the set of newly observed locations for agent \(i\) at time \(t\). The variable \(\mathrm{val}\) indicates whether the location \(x\) is blocked \((\mathrm{val} = 1)\) or free \((\mathrm{val} = 0)\). Through a decentralized approach, the shared map is updated incrementally. Each agent periodically transmits its own \(\Delta m^i_t\) to teammates, who then merge these local changes into their versions of the shared map. This maintains a distributed structure without introducing excessive communication overhead. Formally, the map update is:
\[
M^{(t+1)}
\;=\;
\mathrm{Fuse}\bigl(M^{(t)}, \Delta m^i_t\bigr),
\]
where \(\mathrm{Fuse}(\cdot)\) merges the current global map \(M^{(t)}\) with the newly observed data \(\Delta m^i_t\).

Path blockages often arise when obstacles appear or edges become impassable. If an agent detects a new obstacle or a blocked path, it shares this information with the shared map, and other agents add the update to their local copies of \(M^{(t)}\). Although full synchronization is not essential, most agents will still receive the new data within a practical time frame and adjust their path planning as needed.

\begin{figure}[h]
    \centering
    \includegraphics[height=3in]{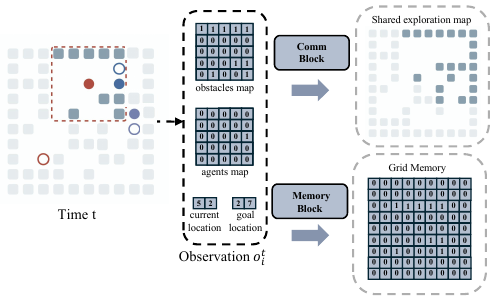}
    \caption{The shared map provides global information that is updated in small increments from local observations.}
    \label{fig:Map}
\end{figure}

Figure~\ref{fig:Map} shows how the global map appears once each agent's observations have been combined. After incorporating newly observed data, an agent can either switch to local RL or continue with a global plan, depending on its strategy. At time \(t\), let \(E^{(t)}\) denote the set of valid edges. If agent \(i\) identifies a blocked edge \(e \in E^{(t)}\), it includes \(e\) in \(\Delta m^i_t\). The updated edge set becomes
\[
E^{(t+1)} 
\;=\; 
E^{(t)} \,\setminus\, \{\,e\},
\]
which removes \(e\) from all future routing. Other agents that receive this update also remove \(e\) from their respective planning graphs. Through periodic exchanges of \(\Delta m^i_t\), each agent maintains an approximate global map. This approach avoids forcing complete synchronization and reduces communication demands while ensuring that agents have timely access to environmental data.

Although the shared map consolidates newly discovered obstacles or blocked paths from multiple agents, each agent still maintains a local memory that stores and processes immediate observations. The memory grid fulfills this need: it remains at the agent level, expands as the agent moves, and accumulates locally relevant data over time. By maintaining the memory grid, each agent can respond more quickly to environmental changes without relying on full global synchronization. This combination of a periodically updated shared map and a locally expanded memory grid supports navigation in dynamic environments by integrating global knowledge and local history.

As shown in Figure~\ref{fig:Map}, when the agent observes changes in environmental obstacle information, it promptly writes these updates into its own grid memory. The specific method is to insert new grid cells to store the newly observed values and expand the grid memory range as the agent moves throughout the environment. This mechanism not only allows the grid memory to accumulate and retain earlier observations but also enables the model to make full use of past information in subsequent path planning and decision-making. Consequently, it greatly improves adaptability and planning efficiency in dynamic environments.

\section{Experiment}

\subsection{Environment Setup and Training Setting}
We use a 4-neighbor grid to represent the environment, where neighboring cells connect horizontally or vertically. Each map cell is either free or blocked, and agents can only move through free cells. If an agent is at coordinate \((i,j)\), its observation range is a square neighborhood with radius \(R=4\), resulting in an observation matrix of size \((2R+1)\times (2R+1)\). This matrix shows local obstacles and nearby agents but does not reveal other agents' goals or planned actions. Consequently, the agent lacks complete knowledge of the environment. Following \cite{sharon2013increasing}, each agent exits the map upon reaching its goal and is no longer involved in collision checks. In settings with many obstacles, an agents lingering at its destination might block pathways. In real deployments, an agent that has completed its main task typically leaves the area or does not hinder other agents.

In the training process, we place 64 agents simultaneously on a randomly generated \(64\times 64\) grid map and allow each agent's policy to run for up to 512 steps. Once an agent reaches its goal, or the total steps exceed 512, the episode ends, and a new \(64\times 64\) grid map is generated for the next round. The obstacle ratio is chosen randomly between 15\% and 45\%.

\subsection{Experimental Results}
\begin{figure}[!h]
    \centering
    \begin{subfigure}[b]{0.45\textwidth}
        \centering
        \includegraphics[width=\textwidth]{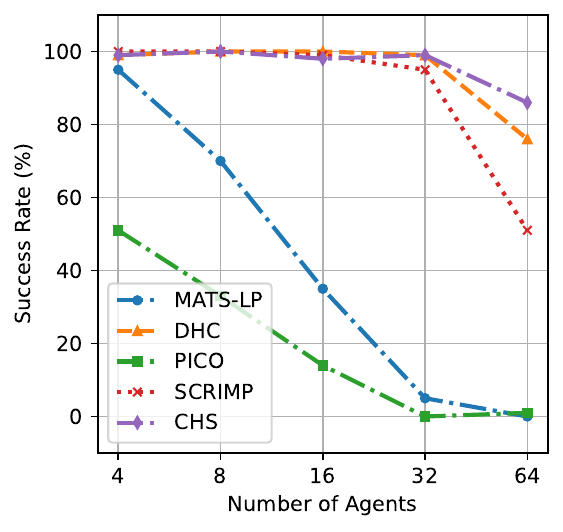}
        \label{fig:20_20success_rates}
    \end{subfigure}
    \hspace{5mm}
    \begin{subfigure}[b]{0.45\textwidth}
        \centering
        \includegraphics[width=\textwidth]{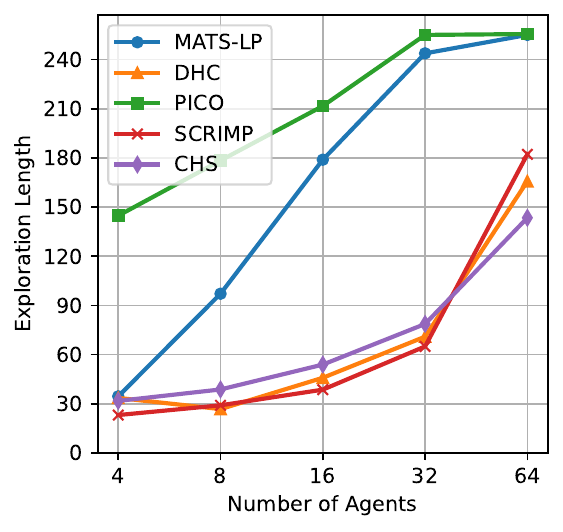}
        \label{fig:20_20ep_length_comparison}
    \end{subfigure}
    \caption{Results on a \(20\times20\) grid map for exploration length and success rate.}
    \label{fig:20*20}
\end{figure}

\begin{table*}[!h]
    \centering
    \scriptsize
    \caption{Comparison of our approach CHS and other algorithms on  \(40\times40\) grid map}
    \label{tab:results}
    \resizebox{1.0\textwidth}{!}{%
    \begin{tabular}{l*{10}{cc}}
        \toprule
        \multicolumn{11}{c}{\(40\times40\) world, 30\% obstacle density} \\
        \cmidrule(lr){2-11}
        Algorithm & \multicolumn{2}{c}{8 agents} & \multicolumn{2}{c}{16 agents} & \multicolumn{2}{c}{32 agents} & \multicolumn{2}{c}{64 agents} & \multicolumn{2}{c}{128 agents} \\
        \cmidrule(lr){2-3}\cmidrule(lr){4-5}\cmidrule(lr){6-7}\cmidrule(lr){8-9}\cmidrule(lr){10-11}
         & SR & EL & SR & EL & SR & EL & SR & EL & SR & EL \\
        \midrule
        MATS-LP & 0.85 & 91.7  & 0.3  & 239.6  & 0.05 & 306.15 & 0    & 319    & 0    & 319 \\
        DHC  & 0.99 & 64.37 & 0.99 & 73.11  & 1    & 83.80  & 0.96 & 127.63 & 0.76 & 221.88 \\
        PICO & 0.05 & 308.52& 0    & 319    & 0    & 319    & 0    & 319    & 0    & 319 \\
        SCRIMP  & 1    & 55.57 & 1    & 62.08  & 1    & 71.95  & 0.94 & 112.83 & 0.51 & 235.94 \\
        CHS & 1    & 85.04 & 0.99 & 95.86  & 0.97 & 105.19 & 0.99 & 131.70 & 0.94 & 207.51 \\
        \midrule
        \multicolumn{11}{c}{\(40\times40\) world, 15\% obstacle density} \\
        \midrule
        MATS-LP & 0.95 & 61.55 & 0.7  & 132.05 & 0.35 & 227.25 & 0.10 & 292.80 & 0    & 319 \\
        DHC  & 0.99 & 59.36 & 1    & 61.92  & 0.99 & 70.17  & 1    & 72.67  & 0.99 & 85.24 \\
        PICO & 0.53 & 184.53& 0.52 & 195.75 & 0.61 & 174.14 & 0.70 & 259.87 & 0.25 & 274.23 \\
        SCRIMP  & 1    & 47.74 & 1    & 52.70  & 1    & 58.33  & 1    & 64.00  & 1    & 74.11 \\
        CHS & 1    & 50.34 & 1    & 55.21  & 1    & 61.74  & 1    & 68.40  & 0.98 & 85.85 \\
        \midrule
        \multicolumn{11}{c}{\(40\times40\) world, 0\% obstacle density} \\
        \midrule
        MATS-LP & 1    & 47.05 & 1    & 52.00 & 1    & 55.80 & 0.45 & 202.40 & 0    & 319 \\
        DHC  & 1    & 47.76 & 1    & 56.14 & 1    & 59.62 & 1    & 64.36  & 1    & 68.87 \\
        PICO & 1    & 48.71 & 1    & 53.49 & 0.99 & 62.80 & 1    & 68.01  & 1    & 78.52 \\
        SCRIMP  & 1    & 45.44 & 1    & 50.71 & 1    & 55.83 & 1    & 60.79  & 1    & 67.82 \\
        CHS & 1    & 45.50 & 1    & 52.73 & 1    & 58.09 & 1    & 64.07  & 1    & 69.13 \\
        \bottomrule
    \end{tabular}%
    }
\end{table*}

In order to examine the scalability and reliability of different algorithms in complex environments, we performed experiments on two grid maps of size \(20\times20\) and \(40\times40\). In both sizes, 100 randomly generated instances ensure the results are not biased by a single scenario. Because the \(40\times40\) grid map covers four times the area of the \(20\times20\)grid map, it greatly increases state space and typical path lengths. Testing different map sizes provides a broader perspective on computational efficiency and solution quality in larger domains. Traditional search-based methods often see a complexity increase as the map grows, whereas a trained learning-based policy can handle various domain sizes more flexibly.

To explore performance under high concurrency, we vary the number of agents from 8 to 64 on the \(20\times20\) grid map and from 8 to 128 on the \(40\times40\) grid map. With more agents, collisions and path conflicts become more frequent, which increases computational costs for conventional planners. A well-structured learning-based approach can mitigate these issues through decentralized decision-making. Incrementally increasing agent population helps reveal each algorithm’s adaptability while controlling for excessive congestion.

For fairness, the maximum number of simulation steps is scaled by map size: 256 steps for the \(20\times20\) grid map and 320 steps for the \(40\times40\) grid map. This ensures agents have adequate time to complete tasks, preventing conditions where limited steps might artificially restrict certain methods. Aligning the step limits with map dimensions is a common practice to keep all methods under consistent conditions.

We compare MATS-LP~\cite{skrynnik2024decentralized}, DHC, SCRIMP, and PICO. DHC, SCRIMP, and PICO use communication-based learning and rely on global obstacle data to construct heuristic channels, whereas MATS-LP utilizes global information for planning. By examining these baselines that require global knowledge, we highlight the advantage of our approach. Since MATS-LP, DHC, SCRIMP, and PICO assume an agent becomes stationary after reaching its goal, we remove finished agents to avoid interference, ensuring a uniform evaluation. Results show that MATS-LP leverages global information to perform well in small or sparse maps, but faces rising computational demands and higher collision rates in larger or denser maps. DHC, SCRIMP, and PICO are partially observable methods that rely on communication to maintain teamwork. Our method also achieves high success under dense conditions with only a small increase in average path length.

Traditional solutions such as MATS-LP do not require long time training but depend heavily on global knowledge, which is effective in small cases \cite{tjiharjadi2022systematic}. However, as agent count or map size grows, computational demands escalate, hindering real-time planning. Learning-based approaches (DHC, SCRIMP, and PICO) train policies in partially observable setups and use decentralized decisions, so their effectiveness depends on training data and generalization. Results confirm that learning-based policies adapt better to large-scale environments, and we compare their success rates and efficiency with more classical methods.

MATS-LP plans independently without sharing agent decisions, while learning-based methods rely on communication for better coordination. For example, DHC broadcasts key information to neighbors, SCRIMP uses a transformer-based structure for global exchanges, and PICO shares crucial data according to priority. Communication can help reduce path conflicts and enhance collaboration but adds overhead. Varying the agent population and scenarios shows how each method’s communication design affects multi-agent path planning.

We first analyze results on the \(20\times20\) grid (Figure~\ref{fig:20*20}), where the agent count ranges from 4 to 64. Since CHS does not rely on complete global knowledge, it achieves performance comparable to DHC and SCRIMP for 4, 8, 16, and 32 agents, while consistently surpassing MATS-LP and PICO. As the agent population grows, performance gaps among these methods narrow. Nevertheless, at 64 agents, CHS attains the highest success rate, confirming its reliability when collision risks escalate.

Turning to the larger \(40\times40\) grid (Table~\ref{tab:results}), CHS remains robust against increased obstacles and agent density. While MATS-LP and PICO experience sharp drops in success under high congestion, DHC and SCRIMP also see reduced rates once the agent count reaches 128. By contrast, CHS maintains a high success rate, indicating that its primarily local-view approach adapts effectively to blocked edges and crowded regions.

Moreover, CHS’s exploration length remains on par with other learning-based methods, suggesting it leverages communication efficiently without drastically inflating path costs. This local-view design with incremental map updates allows CHS to handle congestion more flexibly than methods that rely on fully centralized obstacle data. By sharing only necessary map changes rather than global information, CHS avoids the heavy overhead associated with synchronized planners. Consequently, even under dense and large-scale conditions, agents can respond swiftly to new blockages or conflicts. The relatively small gap in both success rate and path length compared with other methods further demonstrates that CHS effectively utilizes its communication mechanism, positioning it as a strong candidate for multi-agent path planning in congested environments.

\begin{figure}[h]
    \centering
    \begin{subfigure}[b]{0.45\textwidth}
        \centering
        \includegraphics[width=\textwidth]{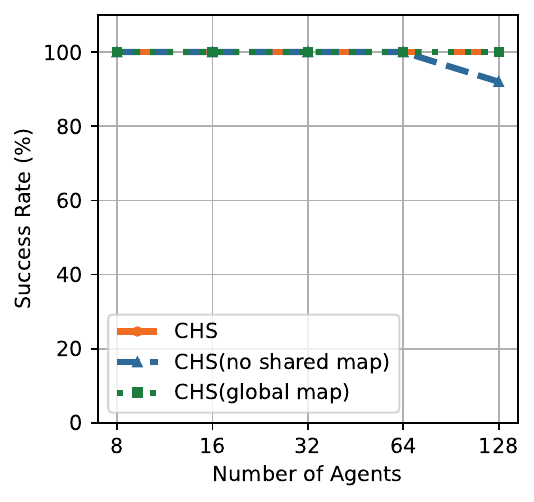}
        \label{fig:ablation1_Success}
    \end{subfigure}
    \hspace{5mm}
    \begin{subfigure}[b]{0.45\textwidth}
        \centering
        \includegraphics[width=\textwidth]{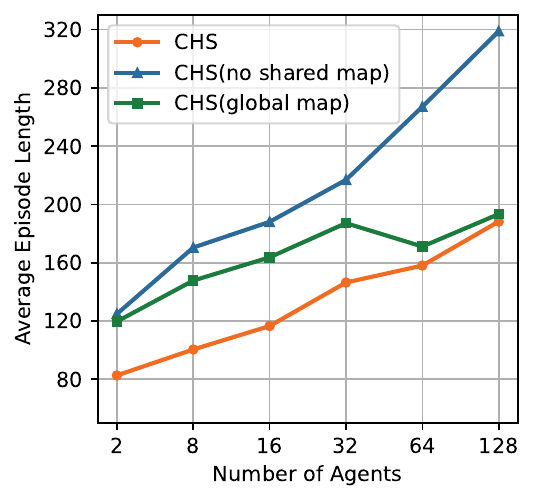}
        \label{fig:ablation1_EL}
    \end{subfigure}
    \caption{Shared Exploration Maps Ablation}
    \label{fig:ablation_combined}
\end{figure}

\subsection{Ablation Study}
To evaluate how each module contributes to overall system performance, we conducted an ablation study focusing on two key components: the shared map mechanism and loop detection. First, we examined the effect of the shared map on an \(80\times80\) grid, as illustrated in Figure~\ref{fig:ablation_combined}, where the obstacle density was 30\% and the number of agents was varied. Each setting was tested on 100 randomly generated maps to ensure consistency. Because each agent can only observe a \(4\times4\) region, using a larger grid highlights how shared mapping influences group-level success.

We tested three configurations: Full Information, in which every agent has complete obstacle data; Shared Map, where agents share locally observed data; and Local Map Only, where agents rely solely on their own observations.
\begin{figure}[!b]
    \centering
    \begin{subfigure}[b]{0.45\textwidth}
        \centering
        \includegraphics[width=\textwidth]{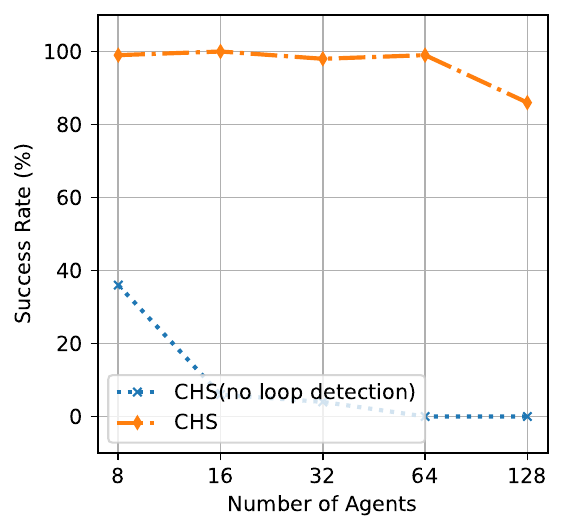}
        \label{fig:ablation2_Success}
    \end{subfigure}
    \hspace{5mm}
    \begin{subfigure}[b]{0.45\textwidth}
        \centering
        \includegraphics[width=\textwidth]{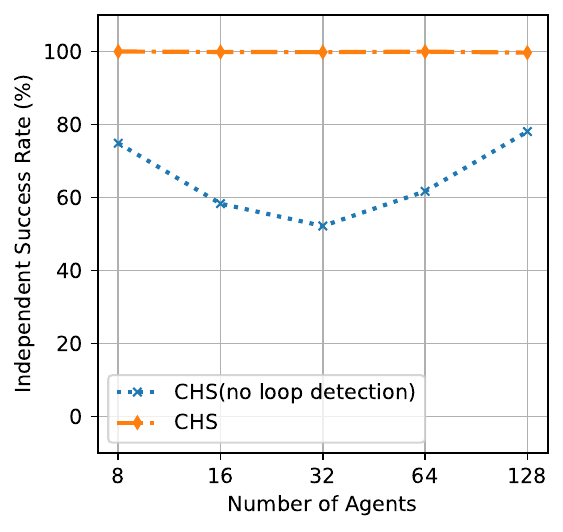}
        \label{fig:ablation2_EL}
    \end{subfigure}
    \caption{Loop Detection Ablation}
    \label{fig:ablation_combined2}
\end{figure}
The results show that removing the shared exploration map substantially lowers success rates. For instance, with 128 agents, a 100\% success rate cannot be achieved without the shared map, whereas Full Information and Shared Map approach full success. Likewise, the average exploration steps remain higher under Local Map Only, indicating that without teammate-supplied data, agents incur higher search costs and reduced success. Meanwhile, comparing Shared Map to Full Information reveals that as agent numbers grow, differences in average steps narrow, highlighting how incremental sharing proves especially beneficial in large-scale multi-agent planning. Completely disabling communication further confirms its importance: success rates drop noticeably, while average steps and collision rates surge.

Next, we investigated the effect of loop detection on a \(40\times40\) grid with 30\% obstacles and 100 randomly generated maps, as depicted in Figure~\ref{fig:ablation_combined2}. Without loop detection, the overall success rate is around 40\% for just 8 agents, and independent completion hovers near 80\%.

\begin{figure}[!h]
    \centering
    \begin{subfigure}{0.47\textwidth}
        \centering
        \includegraphics[width=\textwidth]{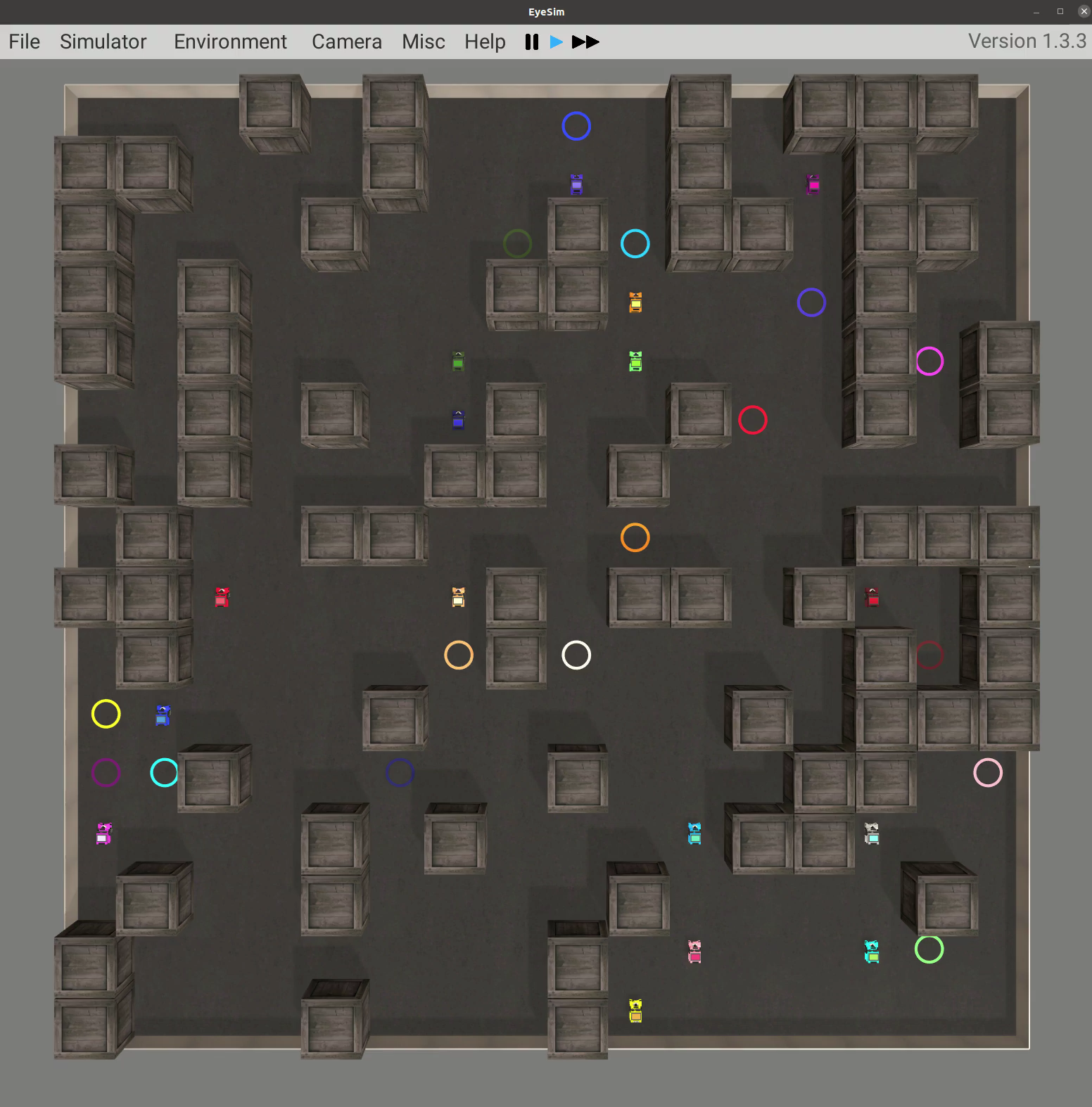}
        \label{fig:Start}
    \end{subfigure}%
    \hspace{0.05\textwidth}
    \begin{subfigure}{0.47\textwidth}
        \centering
        \includegraphics[width=\textwidth]{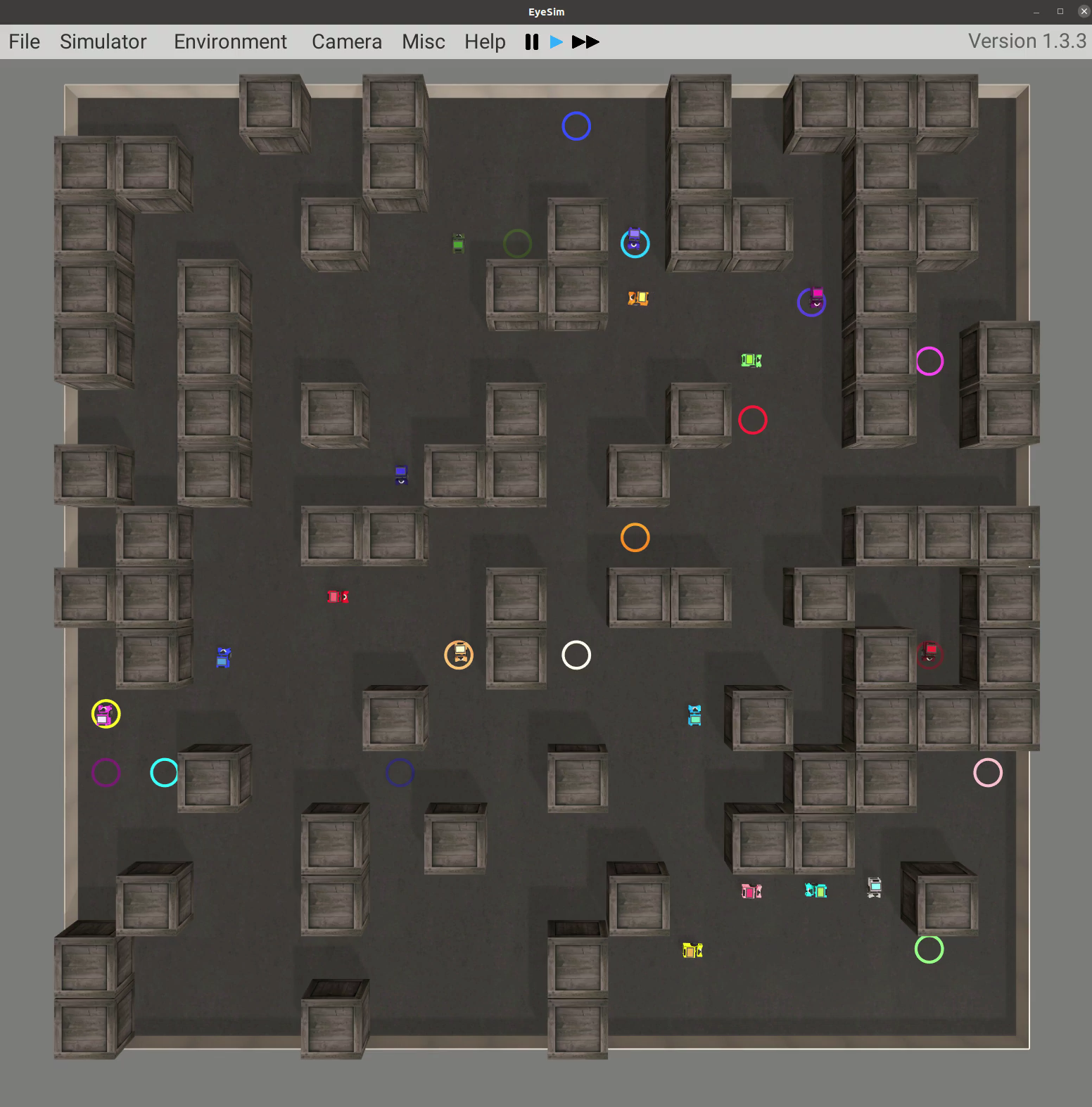}
        \label{fig:Finish2}
    \end{subfigure}\\
    \begin{subfigure}{0.47\textwidth}
        \centering
        \includegraphics[width=\textwidth]{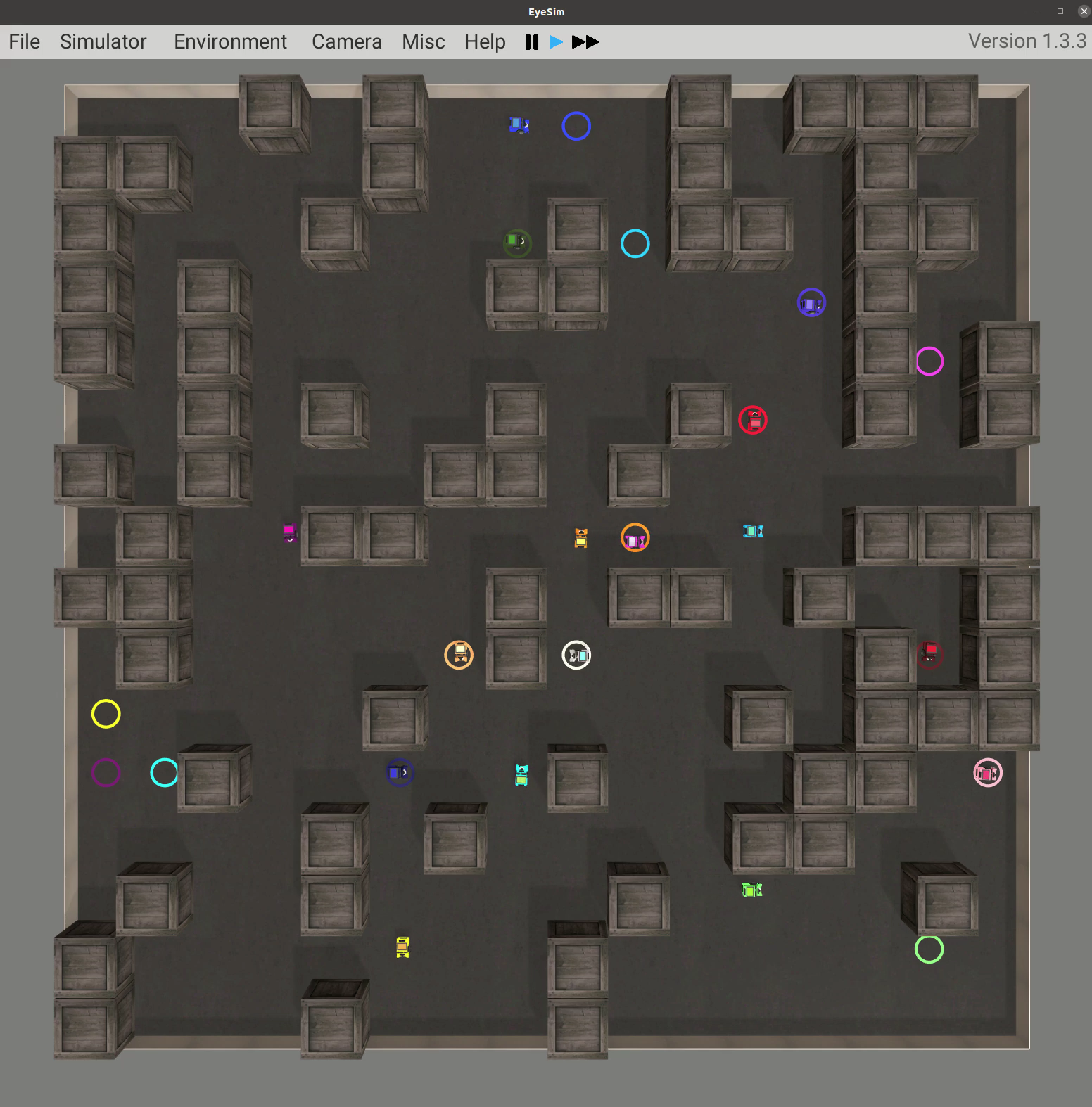}
        \label{fig:Finish9}
    \end{subfigure}%
    \hspace{0.05\textwidth}
    \begin{subfigure}{0.47\textwidth}
        \centering
        \includegraphics[width=\textwidth]{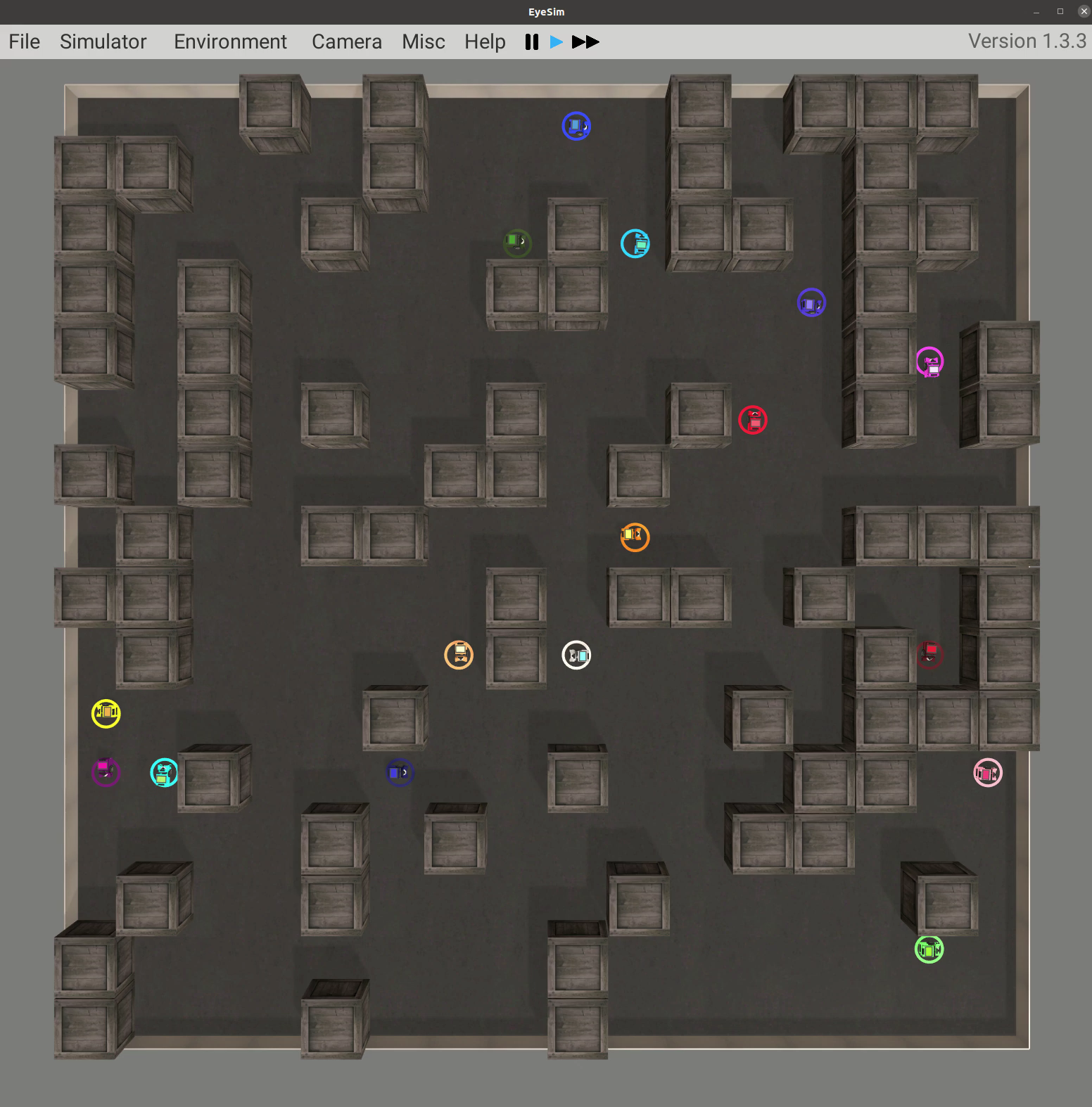}
        \label{fig:Last}
    \end{subfigure}
    \caption{
    1) The initial \(16\times16\) environment with 16 EyeBots and 0.3 obstacle density.
    2) The EyeBots start moving and interacting with the environment as they approach their targets.
    3) A stage when half of the EyeBots have finished.
    4) The final state where all EyeBots reach their assigned positions.
    }
    \label{fig:Eyesim}
\end{figure}

As the agent count increases, success falls to 0, and even independent completion declines sharply by 32 agents, largely due to frequent congestion and cyclical deadlocks in D*Lite.

By contrast, enabling loop detection allows some agents to switch to reinforcement learning more often, avoiding these loops. With loop detection, the independent completion rate stays near 100\% across different agent populations, and only a few rare loops prevent a perfect overall success rate. This outcome confirms that loop detection is crucial for preventing deadlocks and enhancing completion rates.

\subsection{Platform Evaluation}
To gain further insights into the feasibility of our approach in a hardware-oriented context, We used the EyeSim simulator to further test our trained model on 16 robots within a \(16\times16\) maze, as shown in Figure~\ref{fig:Eyesim}. Each robot has a unique color, and a circle of the same color indicates its destination. Once a robot arrives at its goal, it is considered finished.

\begin{figure}[ht]
    \centering
    \begin{subfigure}[b]{0.32\textwidth}
        \centering
        \includegraphics[width=\textwidth]{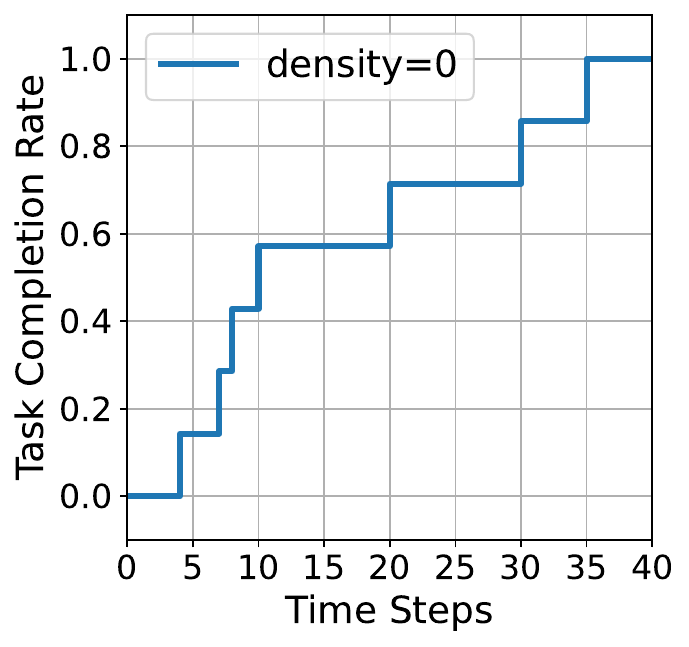}
        \caption{}
        \label{fig:output_step_density0}
    \end{subfigure}
    \hfill
    \begin{subfigure}[b]{0.32\textwidth}
        \centering
        \includegraphics[width=\textwidth]{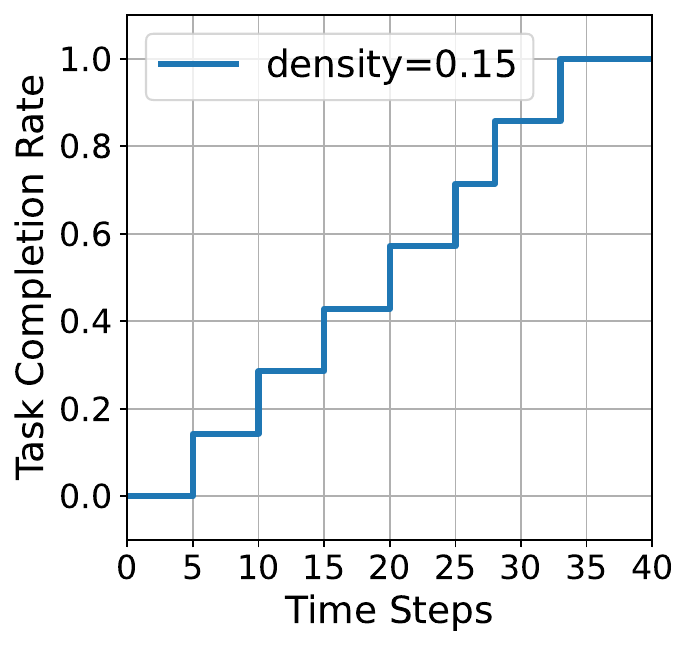}
        \caption{}
        \label{fig:output_step_density0.15}
    \end{subfigure}
    \hfill
    \begin{subfigure}[b]{0.32\textwidth}
        \centering
        \includegraphics[width=\textwidth]{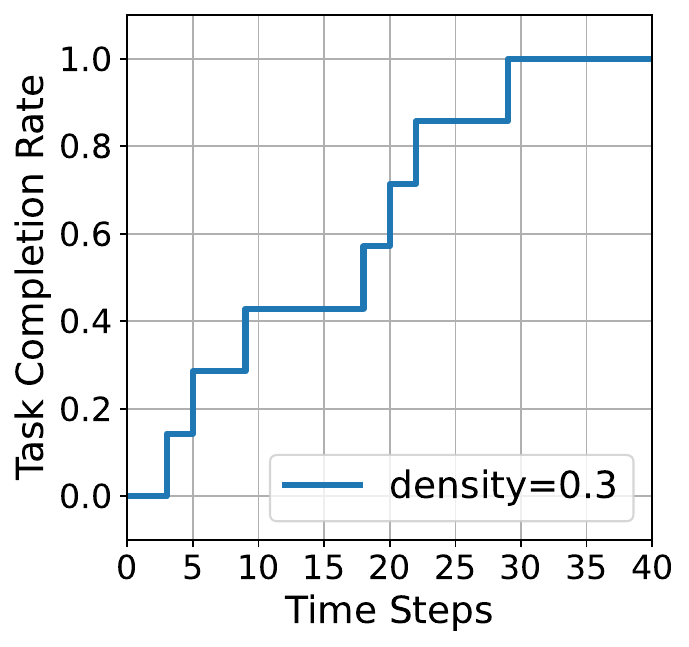}
        \caption{}
        \label{fig:output_step_density0.3}
    \end{subfigure}

    \vspace{2mm}
    \begin{subfigure}[b]{0.32\textwidth}
        \centering
        \includegraphics[width=\textwidth]{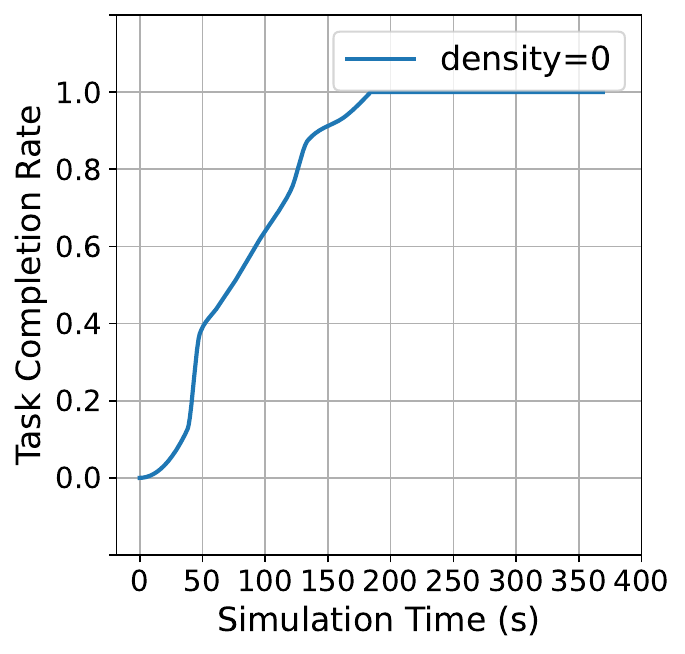}
        \caption{}
        \label{fig:final_plot_pchip_density0}
    \end{subfigure}
    \hfill
    \begin{subfigure}[b]{0.32\textwidth}
        \centering
        \includegraphics[width=\textwidth]{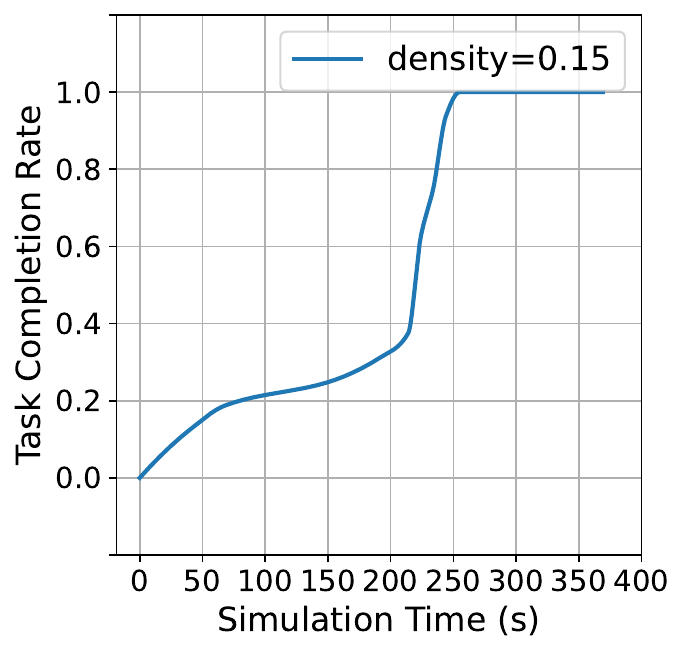}
        \caption{}
        \label{fig:final_plot_pchip_density0.15}
    \end{subfigure}
    \hfill
    \begin{subfigure}[b]{0.32\textwidth}
        \centering
        \includegraphics[width=\textwidth]{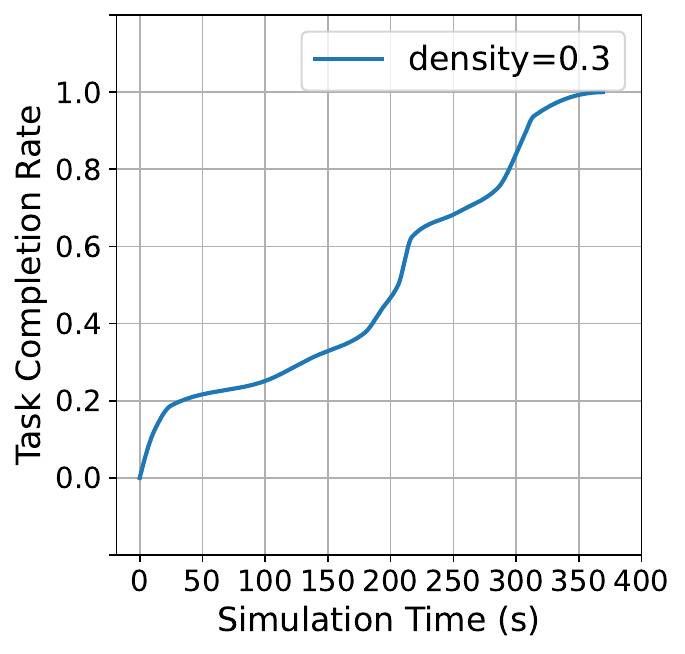}
        \caption{}
        \label{fig:final_plot_pchip_density0.3}
    \end{subfigure}
    \caption{Task completion rate at different obstacle densities, comparing simulation time and discrete time steps.}
    \label{fig:time_diff}
\end{figure}

Figure~\ref{fig:time_diff} illustrates how task completion rate changes over discrete time steps and simulation time under different obstacle densities. Discrete time steps represent the number of planning cycles or algorithmic updates, reflecting computational effort. Simulation time better reflects real-world conditions, as it captures physical movement and communication delays. By comparing these two perspectives, one can see differences between algorithmic progress and real-world operation, which aids in optimizing multi-robot systems for various applications.

As more robots finish their tasks, both discrete time steps and simulation time increase more slowly, and their differences relative to completion rate also decrease. Put simply, when most robots have reached their goals, congestion in the environment drops, so there are fewer idle movements or re-planning cycles. Whether measuring discrete steps or simulation time, the remaining robots can finish more quickly, enhancing overall efficiency.

Early in the process, many robots are still navigating, and path conflicts plus frequent re-planning cause discrete steps and simulation time to rise quickly. As the completion rate climbs (i.e., more robots reach their destinations and stop moving), the number of robots in motion decreases, and collision frequency lessens. Hence, the gaps among discrete steps, simulation time, and completion rate narrow over time. This indicates that, as tasks advance, unnecessary waiting and repeated moves become less likely, allowing resources to be distributed more effectively among remaining robots.

When obstacle density is high, potential paths narrow, making robots more prone to traffic jams or blockages. Heightened path conflict leads to increased re-planning and waiting, which significantly raises discrete time steps and simulation time. Conversely, when there are fewer obstacles, robots can find clearer routes and reduce collision probability, letting them complete tasks faster while also boosting overall success and efficiency.

\section{Conclusion}
In response to the challenge of balancing global optimality and local adaptability for MAPF in incomplete or dynamic environments, this study proposes a hybrid framework that integrates D* Lite with MARL. The system uses D* Lite at the global level to quickly generate feasible reference paths, and applies reinforcement learning at the local level for flexible decision-making based on real-time observations. Through a switching mechanism and a freeze prevention strategy, it coordinates the two methods to avoid deadlock and reduce redundant planning in high-density or highly dynamic scenarios. Experimental results show that the proposed hybrid framework achieves higher arrival rates and better path efficiency. Finally, we deploy the model on a simulation platform to demonstrate that robots can rely on the learned policy to accomplish efficient real-time path planning. Future work will extend this approach to more practical MAPF tasks, such as communication constraints among agents\cite{maoudj2022decentralized,dogrumultitier}, or add dynamic obstacles\cite{,liu2024multi,liu2020mapper} and will investigate the robustness of the system under various restrictive conditions\cite{atzmon2020robust}.

\section*{Declarations}
\setlength{\parindent}{0pt}

\textbf{Author Contributions}
Ning Liu served as the corresponding author, playing a leading role in conceiving, designing, and implementing the research. He also took the lead in drafting the manuscript, ensuring clarity, accuracy, and adherence to academic standards. Sen Shen actively participated in data collection, analysis, and interpretation, and contributed to the writing of the manuscript. Xiangrui Kong contributed to data collection and experimental design, assisted in result analysis, and provided valuable suggestions for improving the manuscript.
Hongtao Zhang was involved in experimental testing and results analysis, and supported the manuscript preparation. Professor Thomas Bräunl provided essential guidance and academic support throughout the research process, especially in system architecture design and methodological refinement, which significantly improved the quality of the research and the final manuscript.

\textbf{Funding}  
The authors declare that no funds, grants, or other support were received during the preparation of this manuscript.

\textbf{Competing Interests}  
The authors have no relevant financial or non-financial interests to disclose.

\section*{Code or Data Availability}
The code that supports the findings of this study is openly available on Github \cite{CHS-MAPF}.

\backmatter

\bibliography{sn-bibliography}

\end{document}